\documentclass[times, preprint, 10pt]{elsarticle}

\usepackage{amssymb}
\usepackage{amsmath}
\usepackage{xspace}
\usepackage{enumitem}
\usepackage{colortbl}
\usepackage[table,xcdraw]{xcolor}
\usepackage{multirow}
\usepackage[normalem]{ulem}
\useunder{\uline}{\ul}{}

\makeatletter
\DeclareRobustCommand\onedot{\futurelet\@let@token\@onedot}
\def\@onedot{\ifx\@let@token.\else.\null\fi\xspace}

\def\eg{\emph{e.g}\onedot}

\def\etal{\emph{et al}\onedot}
\makeatother

\begin{document}

\begin{frontmatter}



\title{Variable Radiance Field for Real-World Category-Specific Reconstruction from Single Image}


\author{Kun Wang$^1$, Zhiqiang Yan$^1$, Zhenyu Zhang$^2$, Xiang Li$^3$, Jun Li$^1$, and Jian Yang$^1$} %% Author name

\address{$^1$PCA Lab, Nanjing University of Science and Technology, China}
\address{$^2$Nanjing University, Suzhou Campus, China}
\address{$^3$Nankai University, China}

%% Abstract
\begin{abstract}
    
%    思路：因为使用了全局的表示，所以需要对齐，强调这一点

Reconstructing category-specific objects using Neural Radiance Field (NeRF) from a single image is a promising yet challenging task. Existing approaches predominantly rely on projection-based feature retrieval to associate 3D points in the radiance field with local image features from the reference image. However, this process is computationally expensive, dependent on known camera intrinsics, and susceptible to occlusions.
To address these limitations, we propose Variable Radiance Field (VRF), a novel framework capable of efficiently reconstructing category-specific objects without requiring known camera intrinsics and demonstrating robustness against occlusions. First, we replace the local feature retrieval with global latent representations, generated through a single feed-forward pass, which improves efficiency and eliminates reliance on camera intrinsics. Second, to tackle coordinate inconsistencies inherent in real-world dataset, we define a canonical space by introducing a learnable, category-specific shape template and explicitly aligning each training object to this template using a learnable 3D transformation. This approach also reduces the complexity of geometry prediction to modeling deformations from the template to individual instances.
Finally, we employ a hyper-network-based method for efficient NeRF creation and enhance the reconstruction performance through a contrastive learning-based pretraining strategy. Evaluations on the CO3D dataset demonstrate that VRF achieves state-of-the-art performance in both reconstruction quality and computational efficiency.

\end{abstract}


%\item Efficient single-image reconstruction method without requiring known camera intrinsic
%\item Resolve coordinate inconsistencies inherent in real-world training data
%\item Deform the category-specific shape template to individual instances
%\item Contrastive learning-based pre-training enhances model performance
%\item Hyper-network-based NeRF creation for efficient instance-specific rendering

%% Keywords

\end{frontmatter}



\section{Introduction}

Creating 3D objects from images is a fundamental problem in computer vision, with applications spanning augmented reality (AR) \cite{ar_2019, TIAN201596}, virtual reality (VR) \cite{vr_2019, bruno20103d}, and animation films \cite{LI2024128487, he2021arch++}, among others. These applications demand efficient, user-friendly methods for generating realistic 3D models.
In recent years, Neural Radiance Field (NeRF) \cite{nerf, WU2025129041, WANG2025129112} has demonstrated significant potential in reconstructing high-quality 3D objects. However, NeRF typically requires numerous input images and extensive instance-specific optimization, making it computationally expensive and less practical for many real-world applications. To address these limitations, various generalizable variants of NeRF \cite{pixelnerf, grf, visionnerf} have been developed to enable object reconstruction from a single image.
These methods generally associate 3D points in the radiance field with local image features from a reference image by projecting the 3D points into 2D space using known camera intrinsics. The retrieved local features are then decoded into density and color values, which are used for volume rendering \cite{volumerender}.

\begin{figure}
    \centering
    \includegraphics[width=0.85\linewidth]{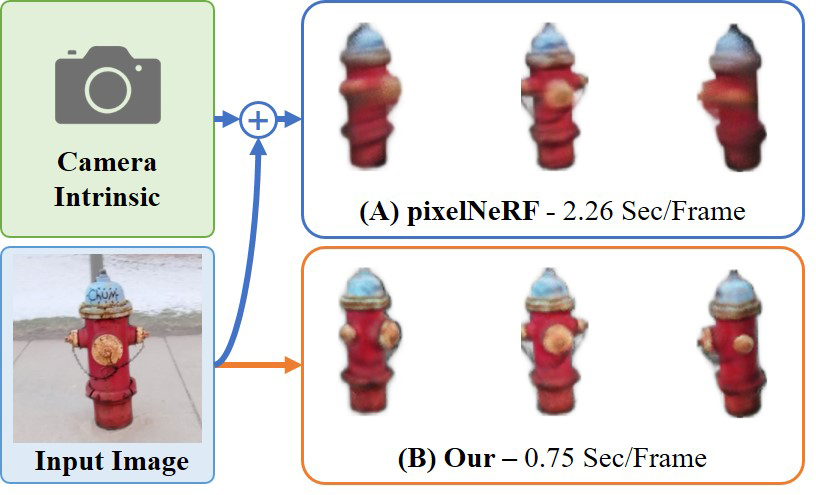}
    \caption{(a) Projection-based feature retrieval correctly locates features from nearby input views but often samples incorrect features from distant views due to occlusions. (b) In real-world datasets, camera poses are individually registered for each instance using SfM methods, resulting in arbitrary coordinate orientations (\eg the blue or green axes) and scales. This leads to coordinate misalignment across different instances.}
    \label{fig.1}
\end{figure}

While these methods can produce high-quality renderings of target views near the reference image, their reliance on known camera parameters and local feature retrieval introduces significant limitations in flexibility and robustness. First, volume rendering requires a large number of queries to determine point properties in the radiance field, leading to frequent feature retrieval and decoding. This process is computationally intensive, resulting in inefficient rendering.
Second, the feature retrieval process depends on accurate camera intrinsics to correctly locate image features. This reliance restricts the applicability of these methods, as precise camera parameters are often unavailable for in-the-wild images. Finally, projection-based feature retrieval is particularly vulnerable to occlusions. As illustrated in Fig. \ref{fig.1} (a), when the target views deviate significantly from the reference view, query points may be incorrectly projected onto surfaces on the opposite side of the object, leading to rendering artifacts.

To address these limitations, we propose Variable Radiance Field, a novel framework for efficiently reconstructing category-specific objects from a single image.
First, we introduce the Object Encoding Module (OEM) to overcome the challenges of projection-based feature retrieval. Specifically, OEM extracts multi-level features from the reference image using a convolutional feature extractor and integrates them into two latent representations that separately encode the object's geometry and appearance. This global encoding approach eliminates the need for local feature retrieval and, consequently, for known camera intrinsics. It also avoids introducing noisy features caused by occlusions and is computationally efficient, as the two representations are generated through a single feed-forward pass. To further enhance the extractor's ability to capture discriminative features, we employ a contrastive learning-based pretraining strategy \cite{simclr,moco}.
Second, we propose the Dynamic Ray Sampling Module (DRSM) to address the coordinate inconsistency in real-world datasets. As shown in Fig. \ref{fig.1} (b), individual camera registration using Structure-from-Motion (SfM) methods results in varying coordinate orientations and scales across instances. This inconsistency significantly degrades the performance of NeRF-based methods, which rely on point positions as input. To resolve this issue, we define a canonical space by introducing a learnable, category-specific shape template and aligning each training object to this template using a learnable 3D transformation. The shape template also serves as a geometry prior for object modeling, reducing the complexity of geometry prediction to modeling deformations from the template to individual instances.
Finally, we propose the Instance Creation Module (ICM) for efficient instance-specific modeling. ICM employs a hyper-network-based \cite{hypernetworks} creation scheme that generates a compact NeRF network for fast, instance-specific rendering, eliminating the need for large, category-specific models during rendering.

In summary, our contributions are as follows:
\begin{itemize}  % [leftmargin=*]
    \item We introduce the OEM, replacing conventional projection-based feature retrieval with a global encoding scheme. This approach eliminates the reliance on known camera intrinsics, enhances computational efficiency, and mitigates noise caused by occlusions.
    \item We develop the DRSM to address coordinate inconsistencies inherent in real-world datasets. By employing a learnable, category-specific shape template, we explicitly align each instance using a learnable 3D transformation, reducing the complexity of geometry prediction to modeling deformations from the template to individual instances.
    \item We propose the ICM, leveraging a hyper-network-based method to generate compact, instance-specific NeRF networks. This design enables fast and efficient rendering for individual objects.
\end{itemize}
Together, these contributions enable our framework to achieve state-of-the-art performance and faster rendering speeds compared to existing methods for the single-image category-specific reconstruction task.

\section{Related Work}

\subsection{Category-Specific Object Reconstruction}

We review related works on reconstructing category-specific objects from 2D images, a well-established and extensively studied task. Early approaches relied on 3D supervision to learn the geometry and appearance of objects, representing them in various forms, such as point clouds \cite{CHEN202146}, meshes \cite{meshrcnn}, voxel grids \cite{3d_r2n2}, and signed distance fields \cite{sal}. Later, methods emerged that eliminated the need for 3D supervision by employing differentiable rendering techniques \cite{ray_consist, mvs_machine}. These methods were primarily trained on synthetic datasets, such as ShapeNet \cite{shapenet}, which feature simpler and more consistent objects than real-world datasets.

To enhance the practical applicability of these methods, several works shifted focus to learning from real-world datasets. For instance, CMR \cite{cmr} proposed a learning framework that recovers the shape, texture, and camera parameters of an object from a single image. U-CMR \cite{ucmr} introduced the camera multiplexing strategy to model distributions over camera poses. Dove \cite{dove} explored methods for learning deformable objects from monocular videos.

In recent years, numerous studies have focused on reconstructing category-specific objects using neural radiance field representations. For example, FiG-NeRF \cite{fignerf} extended NeRF to jointly optimize the 3D representation of object categories and separate objects from their backgrounds. CodeNeRF \cite{codenerf} disentangled shape and texture representations of objects and utilized an auto-decoder architecture \cite{deepsdf} to model unseen objects. LOLNeRF \cite{lolnerf} proposed a method to learn 3D structure from datasets with only single views of each object. However, these methods require independent optimization for each unseen object, limiting their scalability. In contrast, our method can predict the geometry and appearance of unseen objects without the need for independent optimization, improving the reconstruction efficiency.

\subsection{Neural Radiance Field}

Neural Radiance Field (NeRF) \cite{nerf} has emerged as a promising technique for 3D modeling. It encodes a radiance field using a multilayer perceptron (MLP) that takes the point position as input and predicts the radiance and density at that point. While NeRF has demonstrated impressive performance in tasks such as 3D reconstruction and novel view synthesis, it faces limitations when applied to dynamic, unbounded, and large scenes. To address NeRF's limitations in dynamic scenes, D-NeRF \cite{dnerf} introduced an additional time dimension, modeling scene dynamics through deformations in the canonical space. KFD-NeRF \cite{kfdnerf} further integrated NeRF with a Kalman filter-based motion reconstruction framework. For unbounded scenes, NeRF++ \cite{nerfpp} applied separate parameterizations for foreground and background contents, while Mip-NeRF 360 \cite{mipnerf360} introduced a non-linear scene parameterization scheme. To improve efficiency in representing large scenes, Block-NeRF \cite{blocknerf} decomposed a large scene into smaller NeRFs, each trained individually.

NeRF has also been extended to handle in-the-wild images, as demonstrated by NeRF-W \cite{nerfwild}, which generates NeRF representations from real-world images. RawNeRF \cite{rawnerf} adapted NeRF for dark scenes using raw images, while Deblur-NeRF \cite{deblurnerf} improved NeRF's performance in handling blurry images. One of the main limitations of NeRF, however, is its requirement for a large number of images for accurate scene reconstruction. Several methods have sought to address this by optimizing NeRF with fewer images (e.g., 3 or 5 input images), incorporating semantic consistency loss \cite{dietnerf}, depth information \cite{ds-nerf, ddp-nerf}, and geometric constraints \cite{geoaug}.

Despite these advancements, the need for instance-specific optimization and multi-view input images remains a major limitation for NeRF methods. In contrast, our method enables the creation of high-quality NeRF representations for category-specific objects from a single image, without the need for instance-specific optimization, which is more user-friendly and applicable to a broader range of real-world applications.

\subsection{Generalizable Variants of NeRF}

While NeRF has demonstrated impressive 3D modeling capabilities across various tasks, its reliance on independent optimization for each object or scene remains a significant limitation for real-world applications, as the optimization process is time-consuming. To address this issue, several methods have been proposed to predict the 3D properties of unseen objects or scenes without requiring additional optimization. These methods typically rely on projection-based feature retrieval to condition the density and color of points in the radiance field on the input images. Notable examples include pixelNeRF \cite{pixelnerf}, GRF \cite{grf}, IBRNet \cite{ibrnet}, and NerFormer \cite{co3d}. % Limited by the local feature retrieval, these method can only synthesize high-quality novel views near the input images.

To further improve 3D modeling performance, methods like SRF \cite{srf} and MVSNeRF \cite{mvsnerf} incorporated stereo cues from sparse input images to build informative stereo features. GeoNeRF \cite{geonerf} introduced geometry priors using a geometry reasoner and attention-based mechanisms. Point-NeRF \cite{pointnerf} explored a point-based representation to create a neural radiance field more efficiently, while VisionNeRF \cite{visionnerf} employed vision transformers to extract more informative features from the input image.

In contrast to these approaches, our method leverages multi-scale global features to represent the 3D properties of objects. By eliminating the need for local feature retrieval and known camera intrinsics, our framework achieves more efficient and multi-view consistent reconstruction from a single image.

\section{Method}

In this section, we present our framework, termed Variable Radiance Field (VRF). VRF is composed of three key components: the Object Encoding Module (OEM), the Dynamic Ray Sampling Module (DRSM) and the Instance Creation Module (ICM). Before delving into the details of these modules, we first provide a brief overview of the foundational concepts behind Neural Radiance Field (NeRF).

\subsection{Reviewing NeRF}

NeRF encodes a radiance field where each continuous point in the field is characterized by a radiance value $c\in \mathbb{R}^3$ and a density value $\sigma\in \mathbb{R}$, representing color and opacity, respectively. This is achieved using a Multi-Layer Perceptron (MLP), which is an implicit function-based model that takes a 3D point takes a 3D point $\mathrm{p}\in\mathbb{R}^3$ and a unit viewing direction $\mathrm{d}\in\mathbb{R}^3$ as inputs, and predicts the corresponding radiance $c$ and density $\sigma$ for that point: $f_n: (\mathrm{p}, \mathrm{d})\rightarrow (\sigma, c)$. The predicted radiance and density are subsequently used for volume rendering \cite{volumerender} to generate 2D images from arbitrary viewpoints. The rendering equation is given by:

\begin{equation}
    \hat{C}(r)=\int_{t_n}^{t_f}T(t)\sigma(t)c(t)dt,
    \label{eq.render}
\end{equation}
where $\hat{C}(r)$ is the rendered color along a ray $r$, $T(t)=exp(-\int_{t_n}^t\sigma(s)ds)$ handles occlusions and $t_n$ and $t_f$ represent the near and far depth bounds, respectively.
Similarly, the depth map is rendered by computing the mean terminating distance of the ray $r=\mathrm{o}+t\mathrm{d}$, parameterized by the camera origin $\mathrm{o}$ and viewing direction $\mathrm{d}$, via

\begin{equation}
    \hat{D}(r)=\int_{t_n}^{t_f}T(t)\sigma(t)tdt.
    \label{eq.depth}
\end{equation}

The optimization objective for NeRF is to minimize the reconstruction loss, which is computed as the squared difference for all rays $r$:

\begin{equation}
    L_c=\Vert\hat{C}(r)-C(r)\Vert_2,
    \label{eq.mse}
\end{equation}
where $\hat{C}(r)$ and $C(r)$ denotes the rendered pixel and the ground truth color, respectively.

\subsection{Framework Overview}

Given an input image $I$ and its corresponding foreground mask $M$ for a category-specific object, our model can efficiently generate a compact MLP that encodes a radiance field representing the object. Leveraging Eq. \ref{eq.render}, this object can then be rendered from arbitrary viewpoints through volume rendering. To train our model, we require a training dataset $\mathcal{C}=\{(I_i, M_i, E_i, K_i)\}_{i=1}^N$, consisting of $N$ color images $I_i\in \mathbb{R}^{3\times H\times W}$, their foreground masks $M_i\in \mathbb{R}^{1\times H\times W}$, camera poses $E_i\in SE(3)$, and intrinsic parameters$K_i\in \mathbb{R}^{3\times 3}$. While the camera pose $E$ and intrinsics $K$ are necessary during training, they are not required during inference.

\begin{figure}
    \centering
    \includegraphics[width=0.98\linewidth]{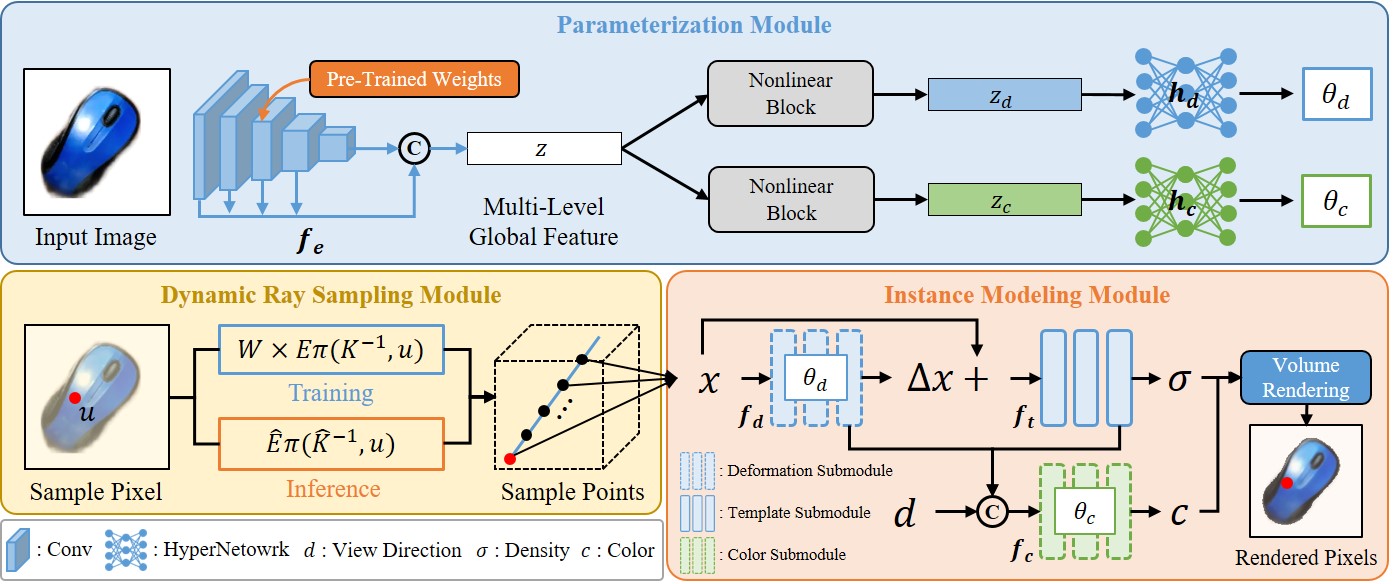}
    \caption{Overall pipeline of our VRF framework. The feature extractor $f_e$ is pre-trained using our contrastive learning-based strategy. $W$ represents the learned transformation that aligns each instance with the template space, while $\pi(\cdot)$ refers to the back-projection operation from the image plane to the camera space. $\hat{E}$ and $\hat{K}$ denote the arbitrary camera pose and intrinsic parameters, respectively.}
    \label{fig.framework}
\end{figure}

The overall framework of VRF is illustrated in Fig. \ref{fig.framework}. It consists of three key components: the Object Encoding Module (OEM), the Dynamic Ray Sampling Module (DRSM) and the Instance Creation Module (ICM). The OEM encodes the object using two global representations that parameterize its appearance and geometry. The DRSM aligns each instance with a canonical template through a learned 3D transformation during training and samples rays using arbitrary camera poses and intrinsics during inference. The ICM employs a hyper-network-based mechanism to generate an instance-specific NeRF representation for efficient rendering. In the following sections, we provide a detailed explanation of each component.

\subsection{Object Encoding Module}

The object encoding module $\Phi_o$ takes an image $I$ and its foreground mask $M$ as input, and generate two latent vectors $z_d$ and $z_c$, which parameterize the geometry and appearance of the object, respectively. This process is mathematically formulated as:
\begin{equation}
    (z_d, z_c)=\Phi_o(I, M).
\end{equation}

To extract object properties from the input image, we use a convolutional feature extractor $f_e$ that outputs multi-level features $z=f_e(I,M)$ at each scale. The features at different scales capture information at varying levels of semantic abstraction, therefore we interpolate them to the same resolution and then concatenate them to form a comprehensive representation of the object properties. We then project this concatenated feature vector $z$ into two distinct latent vectors, $z_d$ and $z_c$, via two nonlinear blocks consisting of linear layers and an intermediate LeakyReLU activation.

\begin{figure}
    \centering
    \includegraphics[width=0.6\linewidth]{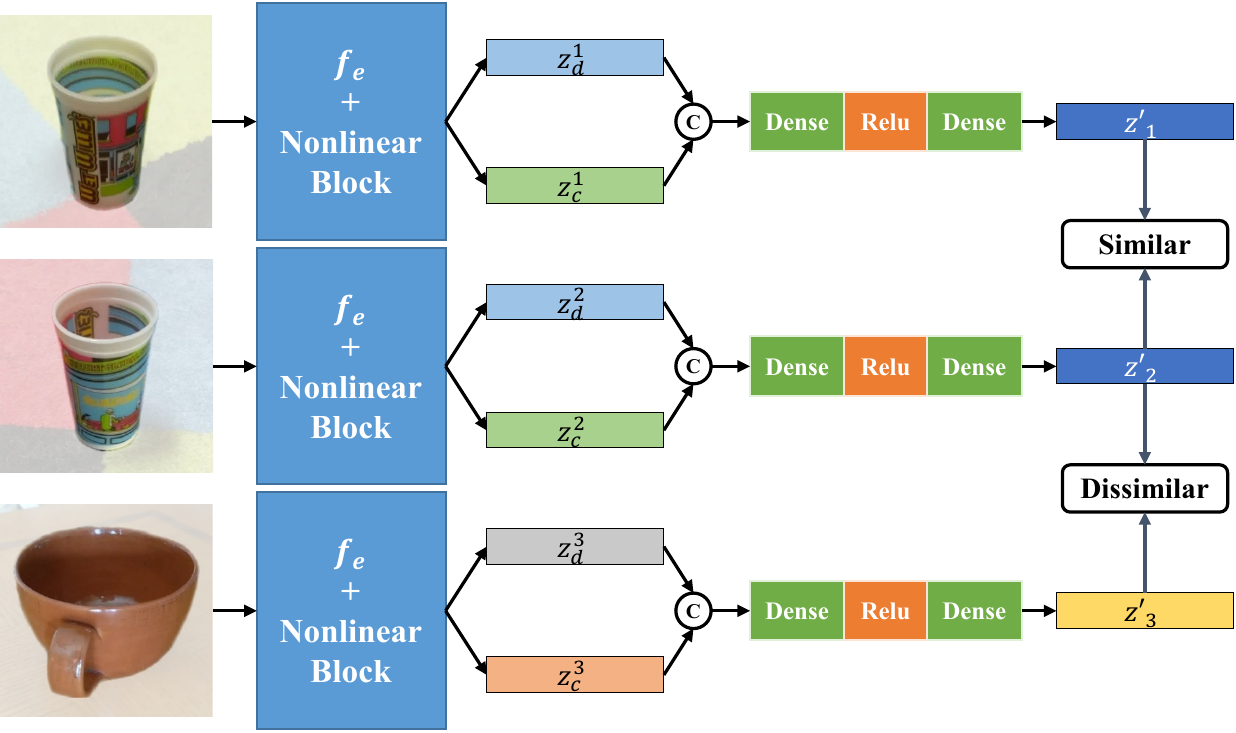}
    \caption{Illustration of the contrastive learning-based pre-training strategy. The feature extractor $f_e$ is trained to generate similar representations for images from the same object instance and dissimilar representations for images from different instances.}
    \label{fig.cl}
\end{figure}

\begin{figure}
    \centering
    \includegraphics[width=0.5\linewidth]{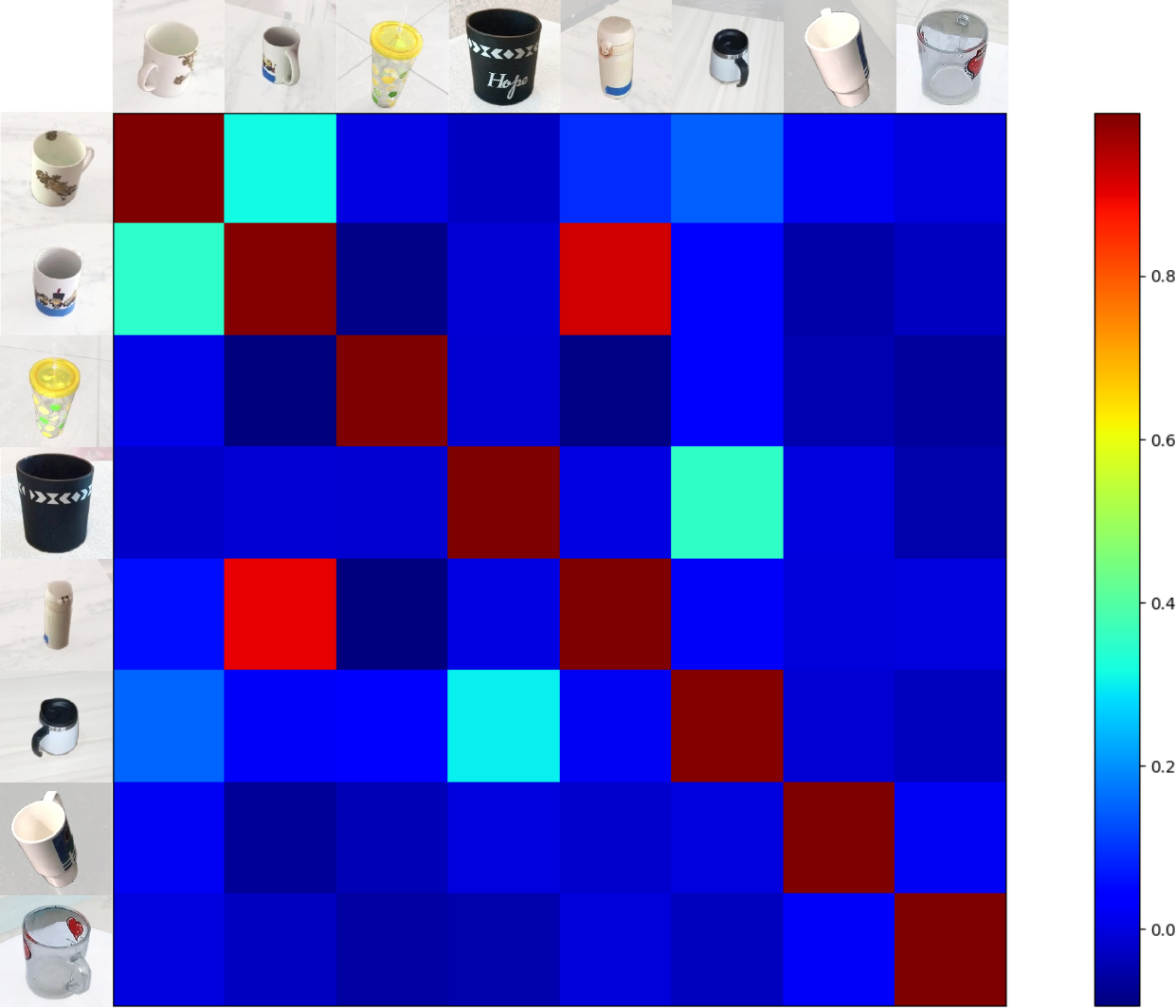}
    \caption{Visual comparison of the similarity of $z'$ across different image pairs. After pre-training, the feature extractor $f_e$ generates similar representations for different input images of the same instance, while producing dissimilar representations for images of different instances.}
    \label{fig.cl_result}
\end{figure}

To improve the feature extractor’s ability to capture object properties, we introduce a contrastive learning-based pretraining strategy. The pretraining scheme is illustrated in Fig. \ref{fig.cl}. The key idea is to ensure that the latent vectors $z_d$ and $z_c$ generated from different views of the same object are similar, while those from different objects are dissimilar. During training, we randomly select different instances within the same object category and sample two images from each instance, each taken from a different viewpoint. These images are fed into the feature extractor $f_e$, which generates the corresponding latent vectors $z_d$ and $z_c$. We then concatenate these vectors and project them into a low-dimensional representation $z'$. To enforce the desired similarity, we use the NT-Xent loss \cite{simclr} to guide the training process:
\begin{equation}
    L_{nt}=-\text{log} \frac{\text{exp}(sim(z'_i,z'_j)/\gamma)}{\sum_{k=1}^{2N}\mathbb{I}_{[k\ne i]}\text{exp}(sim(z'_i,z'_k) / \gamma)},
\end{equation}
where $(z'_i, z'_j)$ are latent vectors generated from different images of the same object, $\gamma$ is a temperature coefficient and $N$ denotes the number of sampled instances. $\mathbb{I}$ is an indicator function, and $sim(\cdot)$ is the similarity function, for which we use cosine similarity to measure the similarity between two vectors. As demonstrated in Fig. \ref{fig.cl_result}, after pre-training, the feature extractor $f_e$ can effectively distinguish between images from different views of the same instance and images from different instances. This confirms that $f_e$ has learned to extract discriminative features, enabling it to better capture object properties, which in turn facilitates more accurate object reconstruction.

\subsection{Dynamic Ray Sampling Module}

The dynamic ray sampling module $\Phi_s$ maps a pixel location $u$ to a ray $r$ that originates from a camera position $o$ and passes through the pixel $u$ with direction $d$. This mapping relies on camera poses $E$ and intrinsic parameters $K$, which can be mathematically represented as:

\begin{equation}
    (o,d)=\Phi_s(u, E, K).
\end{equation}

For real-world datasets, the camera parameters $E$ and $K$ are typically calibrated using Structure-from-Motion (SfM) methods such as COLMAP \cite{colmap_sfm}. However, as each instance is processed individually, the camera poses are registered with different coordinate orientations and scales across instances. As shown in our ablation study, neglecting this misalignment results in significant degradation in reconstruction quality, as NeRF takes point positions as input, and the same coordinates across instances may point to different regions of the objects.

\begin{figure}
    \centering
    \includegraphics[width=0.6\linewidth]{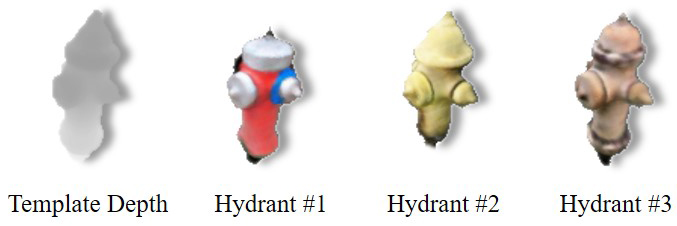}
    \caption{Illustration of instance alignment. We render three different hydrant instances using the same camera pose and intrinsic parameters. All instances are correctly aligned with the predefined template. The template depth is rendered using Eq. \ref{eq.depth}.}
    \label{fig.align}
\end{figure}

To address this challenge, we define a canonical space by introducing a category-specific shape template and optimize a 3D similarity transformation $W \in S(3)$ to align each instance with this template. The shape template is initialized by optimizing a vanilla NeRF MLP $f_t$ with a randomly selected instance from the training data. The radiance output layers in $f_t$ are discarded. During training, the shape template is fine-tuned to reflect the average shape of the category, while individual instance shapes are generated by deforming the template. By limiting the deformation, we ensure that each instance aligns with the template, as shown in Fig. \ref{fig.align}. The ray origin $o$ and direction $d$ at the training phase are computed as:

\begin{equation}
    (o,d)\leftarrow W\times E\pi(K^{-1}, u),
\end{equation}
where $\pi(\cdot, \cdot)$ denotes the back-projection, and $\times$ represents matrix multiplication.

Since each instance is aligned with the canonical template during training, all instances share a common coordinate system. During inference, the transformations $W$, camera poses $E$, and intrinsic parameters $K$ are no longer needed. We can now render an object from any viewpoint using an arbitrarily given camera pose $\hat{E}$ and intrinsic parameters $\hat{K}$. The ray origin $o$ and direction $d$ during inference are computed as:

\begin{equation}
    (o,d)\leftarrow \hat{E}\pi(\hat{K}^{-1}, u).
\end{equation}

\subsection{Instance Creation Module}

The instance creation module generates a compact NeRF network for efficient instance-specific rendering. This compact network consists of three submodules: a deformation submodule $f_d$, a template submodule $f_t$, and a color submodule $f_c$. The $f_t$ corresponds to the shape template, as discussed earlier, while the network weights of $f_d$ and $f_c$ are generated by two hyper-networks $h_d$ and $h_c$, respectively.

The shape template not only defines a canonical space for the instances but also reduces the complexity of geometry modeling by representing deformations from the template to individual instances. Since instances within the same category tend to share similar geometries, this approach introduces informative geometry priors. To predict instance-specific density $\sigma$ from the input image, we employ a hyper-network $h_d$ that takes the latent representation $z_d$ as input and generates the deformation submodule $f_d$ with weights $\Theta_d=h_d(z_d)$. The deformation submodule $f_d$ predicts the instance-specific deformation $\Delta x=f_d(x, \Theta_d)$ relative to the template $f_t$, and the instance-specific density at a point $x$ is then predicted as:

\begin{equation}
    \sigma=f_t(x+f_d(x, \Theta_d)).
\end{equation}
The color $c$ for point $x$ is predicted by the color submodule $f_c$, with weights $\Theta_c=h_c(z_c)$ generated by the hyper-network $h_c$ from the latent representation $z_c$. The input to $f_c$ consists of the view direction $d$ and feature maps $F_d$ and $F_t$, which are the outputs from the last layers of the deformation and template submodules, respectively. The view direction $d$ accounts for view-dependent appearance, while the feature maps $F_d$ and $F_t$ provide additional semantic information. The color $c$ at point $x$ is predicted as:

\begin{equation}
    c=f_c(F_d,F_t,d).
\end{equation}
Note that the weights $\Theta_d$ and $\Theta_c$ are generated in a single feed-forward pass, while the frequent queries for instance-specific density $\sigma$ and color $c$ during volume rendering are handled by the compact NeRF, which significantly improves rendering efficiency.

\subsection{Loss Function}

We use a weighted version of Eq. \ref{eq.mse} to measure the difference between the rendered pixel $\hat{C}(r)$ and the ground truth pixel $C(r)$ for each ray $r$:

\begin{equation}
    L_p = w_t \cdot \Vert \hat{C}(r) - C(r) \Vert^2,
\end{equation}
where $w_t$ is a weight term based on the viewpoint difference between $r$ and the input image, computed using cosine similarity and normalized by the Softmax function over a batch of sampled rays. We assign higher weight to rays from near viewpoints, as rays from distant viewpoints are less predictable.

Next, we apply a foreground masking loss to enforce the density of points beyond the object to be zero. Specifically, we define a binary mask $\hat{M} = [M > 0.5]$ using the Iverson bracket $[\cdot]$ to mark foreground pixels. A weighted $\ell_1$ loss is applied to ensure consistency between the ray opacity $\hat{O}(r)$ and the corresponding mask $\hat{M}(r)$:

\begin{equation}
    L_f = w_f \cdot \Vert \hat{O}(r) - \hat{M} \Vert_1, 
\end{equation}
where we use $w_f = \vert M - 0.5 \vert$ to measure the confidence of $\hat{M}(r)$, and the ray opacity $\hat{O}(r)$ is computed as:

\begin{equation}
    \hat{O}(r) = \int_{t_n}^{t_f} T(t) \sigma(t) , dt.
\end{equation}

To ensure that the shape template learns meaningful category geometry and that the transformation $W$ properly aligns each instance, we apply an $\ell_2$ regularization to penalize large deformations:
\begin{equation}
    L_r=\Vert \Delta x\Vert_2^2.
\end{equation}
Additionally, we use the point pair regularization \cite{template} to prevent extreme distortions:
\begin{equation}
    L_d=\sum_{i\ne j}\mathop{\max}\left(\frac{\Vert\Delta x_i-\Delta x_j\Vert_2}{\Vert x_i-x_j\Vert_2}-\epsilon, 0\right),
\end{equation}
where $\epsilon$ is a parameter that controls the distortion tolerance.

Finally, the total loss is the weighted sum of the individual losses:
\begin{equation}
    L=L_p+\lambda_fL_f+\lambda_rL_r+\lambda_dL_d,
\end{equation}
where $\lambda_f$, $\lambda_r$, and $\lambda_d$ are weight coefficients, and are set to 0.4, 0.1, and 0.1, respectively, in our experiments.

\section{Experiment}

We conduct experiments to evaluate the performance and effectiveness of our method for single-image category reconstruction. We compare our approach with existing state-of-the-art (SOTA) methods, demonstrating superior reconstruction quality and faster inference speed. Additionally, we perform ablation studies to analyze the contribution of each component in our framework. In the following sections, we first outline the experimental setup and implementation details, followed by a presentation of the results and corresponding analysis.

\subsection{Experimental Setup}

We conduct our experiments on five categories from the CO3D dataset \cite{co3d}: banana, cellphone, mouse, hydrant, and cup. For consistency, we follow the official data split for training and testing. Since our objective is to reconstruct objects from a single image, we use only the first image from each test sample as input and render the target image from the specified viewpoint. We compare our framework with five state-of-the-art methods: pixelNeRF \cite{pixelnerf}, Henzler \etal \cite{un3d}, GRF \cite{grf}, NerFormer \cite{co3d}, and VisionNeRF \cite{visionnerf}. Each method is trained separately on each category.

To evaluate reconstruction quality, we use three metrics: PSNR, SSIM \cite{ssim}, and LPIPS \cite{lpips}. PSNR measures pixel-wise error and favors mean color prediction, while SSIM and LPIPS assess structural similarity and perceptual agreement, respectively. Note that PSNR is computed only over the foreground pixels, as masked by $M$. Additionally, we measure the average rendering time across different methods by rendering 100 images at a resolution of $128 \times 128$ on a single TITAN RTX GPU.

\begin{table}[t]
    \centering
    \small
    \caption{Quantitative comparison with existing methods on reconstruction quality and rendering speed. The best result is in \textbf{bold} and the second is \uline{underlined}.}
    \label{tab.compare}
    \resizebox{0.98\linewidth}{!}{
        \begin{tabular}{cc||ccccc||c}
            \hline
            %            \multicolumn{2}{c||}{Metric} & PixelNeRF \cite{pixelnerf} & Henzler \etal \cite{un3d} & GRF \cite{grf} & NerFormer \cite{co3d} & VisionNeRF \cite{visionnerf} & Our \\ \hline
            \multicolumn{2}{c||}{Metric} & pixelNeRF & Henzler \etal & GRF & NerFormer & VisionNeRF  & Our \\ \hline
            \multicolumn{1}{c||}{\multirow{3}{*}{\rotatebox{90}{Banana}}}    & PSNR $\uparrow$    & 20.161 & 18.574      & {\ul 22.858} & 19.170         & 12.086 & \textbf{22.987} \\
            \multicolumn{1}{c||}{}                                           & SSIM $\uparrow$    & 0.503  & 0.813       & 0.688        & {\ul 0.822}    & 0.772  & \textbf{0.861}  \\
            \multicolumn{1}{c||}{}                                           & LPIPS $\downarrow$ & 0.401  & 0.201       & 0.310        & {\ul 0.192}    & 0.269  & \textbf{0.168}  \\ \hline
            \multicolumn{1}{c||}{\multirow{3}{*}{\rotatebox{90}{Mouse}}}     & PSNR $\uparrow$    & 18.696 & 14.904      & {\ul 21.029} & 17.017         & 11.852 & \textbf{22.281} \\
            \multicolumn{1}{c||}{}                                           & SSIM $\uparrow$    & 0.395  & 0.561       & 0.563        & {\ul 0.666}    & 0.473  & \textbf{0.769}  \\
            \multicolumn{1}{c||}{}                                           & LPIPS $\downarrow$ & 0.474  & 0.396       & 0.398        & {\ul 0.308}    & 0.421  & \textbf{0.240}  \\ \hline
            \multicolumn{1}{c||}{\multirow{3}{*}{\rotatebox{90}{Hydrant}}}   & PSNR $\uparrow$    & 15.773 & 15.040      & {\ul 17.670} & 15.927         & 10.216 & \textbf{19.009} \\
            \multicolumn{1}{c||}{}                                           & SSIM $\uparrow$    & 0.605  & 0.775       & {\ul 0.788}  & 0.764          & 0.641  & \textbf{0.847}  \\
            \multicolumn{1}{c||}{}                                           & LPIPS $\downarrow$ & 0.346  & {\ul 0.205} & 0.235        & 0.213          & 0.298  & \textbf{0.162}  \\ \hline
            \multicolumn{1}{c||}{\multirow{3}{*}{\rotatebox{90}{Cellphone}}} & PSNR $\uparrow$    & 16.882 & 10.458      & {\ul 19.673} & 13.691         & 9.991  & \textbf{20.945} \\
            \multicolumn{1}{c||}{}                                           & SSIM $\uparrow$    & 0.460  & 0.583       & 0.644        & {\ul 0.705}    & 0.532  & \textbf{0.841}  \\
            \multicolumn{1}{c||}{}                                           & LPIPS $\downarrow$ & 0.432  & 0.375       & 0.353        & {\ul 0.281}    & 0.401  & \textbf{0.159}  \\ \hline
            \multicolumn{1}{c||}{\multirow{3}{*}{\rotatebox{90}{Cup}}}       & PSNR $\uparrow$    & 16.976 & 15.371      & {\ul 19.387} & 16.611         & 10.196 & \textbf{20.903} \\
            \multicolumn{1}{c||}{}                                           & SSIM $\uparrow$    & 0.381  & 0.656       & 0.556        & \textbf{0.696} & 0.488  & {\ul 0.684}     \\
            \multicolumn{1}{c||}{}                                           & LPIPS $\downarrow$ & 0.521  & {\ul 0.324} & 0.418        & \textbf{0.298} & 0.430  & 0.333           \\ \hline
            \multicolumn{2}{c||}{Sec / Frame}                                                     & 2.26   & {\ul 0.96}        & 1.65         & 1.05           & 2.19   & \textbf{0.75}   \\ \hline
        \end{tabular}
    }
\end{table}

\subsection{Implementation Detail}

We implement our framework using the PyTorch \cite{pytorch} and PyTorch3D \cite{pytorch3d} libraries. For the feature extractor $f_e$, we use a ResNet-18 \cite{resnet} architecture, removing the batch normalization \cite{batchnorm} layers, as we found them to be unstable when processing images containing only the foreground. The dimensions of the latent vectors $z_d$ and $z_c$ are set to 512.
The 3D similarity transformation $W$ is parameterized using a 10-dimensional vector: 6 elements for rotation, 3 for translation, and 1 for scaling. The rotation is represented using the continuous representation proposed in \cite{rotation}.

In the instance creation module, we use MLPs with 192 hidden neurons and ELU \cite{elu} nonlinearities for all three submodules. The submodules $f_d$, $f_t$, and $f_c$ consist of 6, 8, and 3 linear layers, respectively. The hypernetworks $h_d$ and $h_c$ generate weights for each linear layer in $f_d$ and $f_c$ using a sub-hypernetwork, which is a 3-layer MLP. The first two layers of the sub-hypernetwork have 512 and 48 neurons, respectively.

For training, we optimize each model for a specific object category using the Adam \cite{adam} optimizer for 1200 epochs. Different learning rates are used for various components: $5 \times 10^{-5}$ for $f_t$, $1 \times 10^{-2}$ for $W$, $5 \times 10^{-4}$ for the hypernetworks, and $1 \times 10^{-4}$ for the remaining components. The learning rate is halved every 400 epochs.

\begin{figure}[t]
    \centering
    \includegraphics[width=0.98\linewidth]{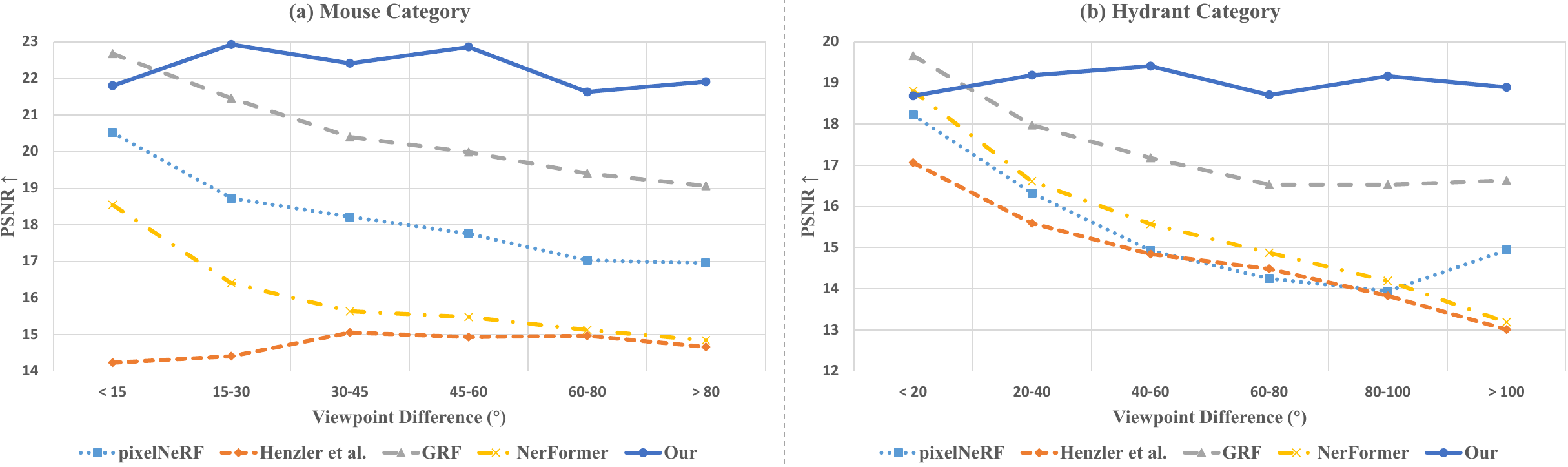}
    \caption{Comparison with existing methods on (a) Mouse and (b) Hydrant categories. We report the PSNR metric at different levels of viewpoint difference (°) between input and target images.}
    \label{fig.lines}
\end{figure}

\subsection{Comparison with Existing Methods}

We present the quantitative results comparing our method with existing approaches in terms of reconstruction quality and rendering speed in Table \ref{tab.compare}. As shown, our framework outperforms all existing methods in both reconstruction quality and rendering speed. Our method achieves the highest PSNR metric across all categories. For example, on the Hydrant category, we observe a $7.58\%$ improvement over GRF, the second-best method. Additionally, our method excels in SSIM and LPIPS metrics for most categories. Specifically, on the Cellphone category, we improve the SSIM and LPIPS metrics by $19.3\%$ and $43.4\%$, respectively, compared to NerFormer, the second-best method. These results highlight the effectiveness of our method for category-specific object reconstruction task.

Our method also achieves the faster rendering speed. Compared to the second-fastest method, Henzler \etal, our method achieves a $21.9\%$ faster rendering speed while improving the average PSNR metric across all categories by $42.7\%$. This performance gain is due to our hyper-network-based mechanism in the instance creation module, which generates a compact MLP for instance-specific rendering, significantly improving the rendering efficiency.

\begin{figure}[t]
    \centering
    \includegraphics[width=\linewidth]{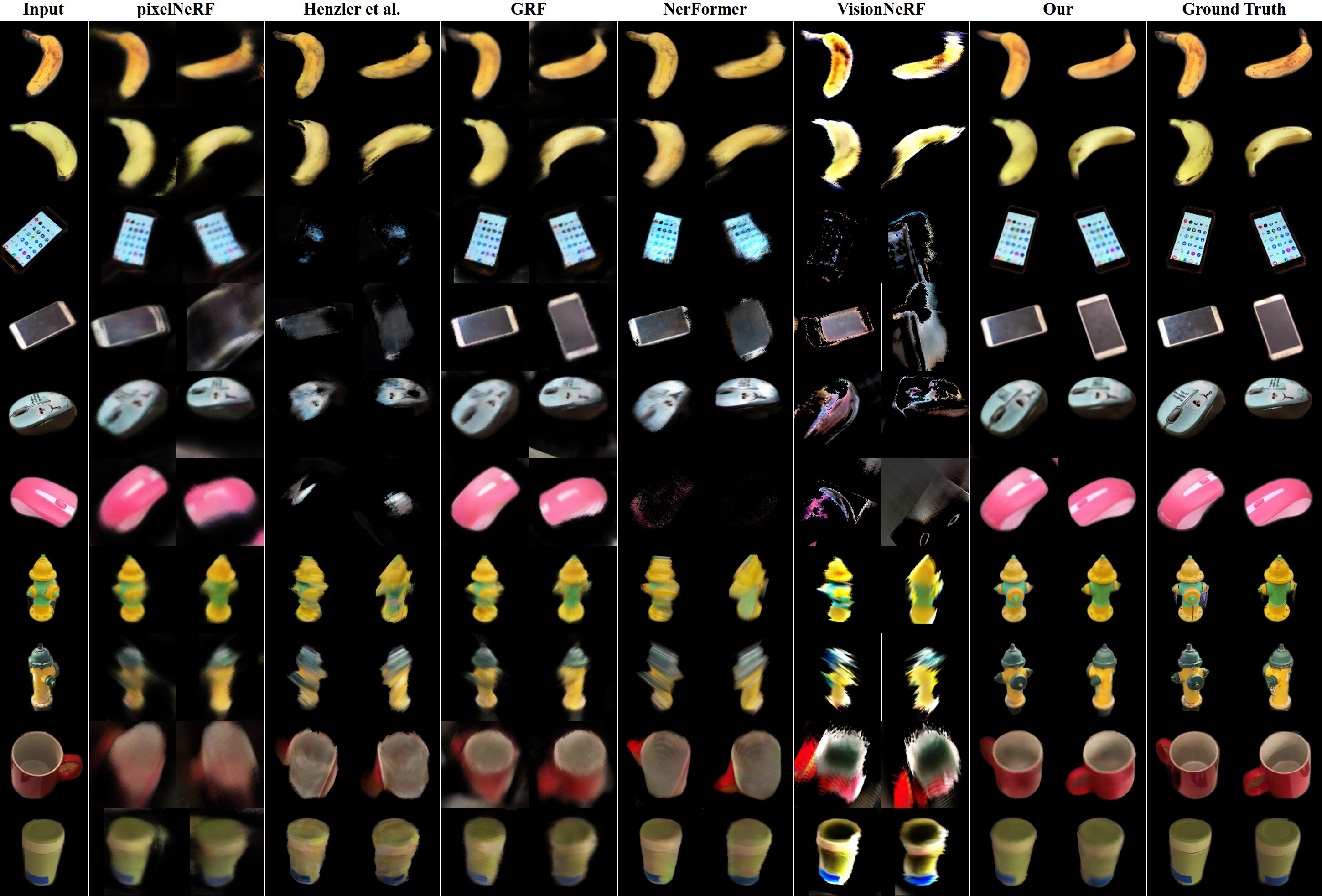}
    \caption{Qualitative comparison with existing approaches. We reconstruct each instance using the input image and render two novel views for each instance.}
    \label{fig.compare}
\end{figure}

To evaluate the robustness of our method against variations in input image viewpoints, we categorize test samples based on the viewpoint difference between the input and target images, and compute the average PSNR for each group. The results are presented in Fig. \ref{fig.lines}. As the viewpoint difference increases, the performance of competing methods deteriorates significantly across both categories. For instance, the PSNR of GRF decreases by $17.4\%$ in the Mouse category, dropping from "$<15$" to "$>80$". In contrast, our method maintains consistent performance across all groups in both categories, highlighting its superior robustness to viewpoint variations.

\begin{figure}
    \centering
    \includegraphics[width=0.6\linewidth]{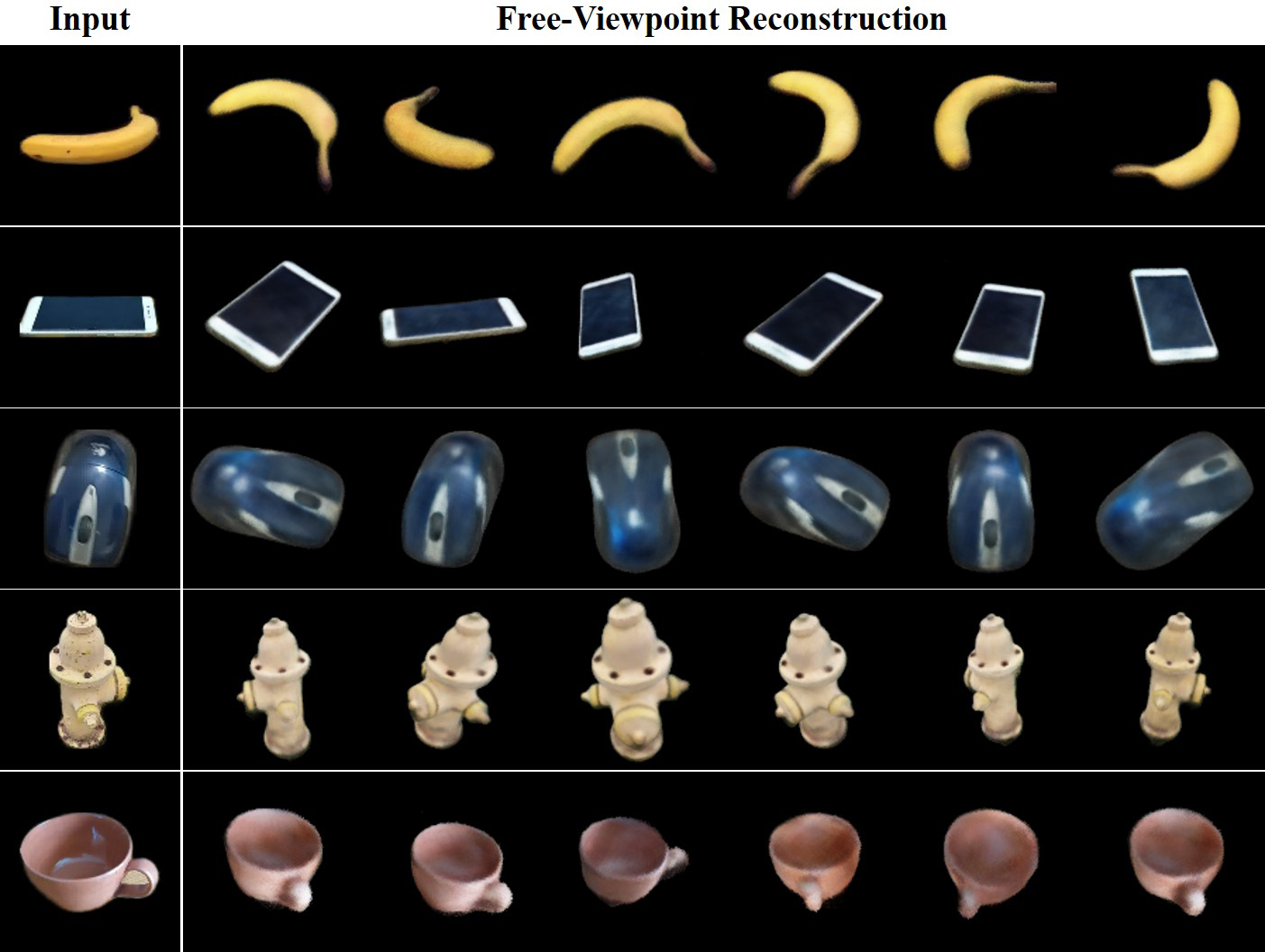}
    \caption{Free-viewpoint rendering result. We reconstruct the objects using the input images and render novel views under six random viewpoints.}
    \label{fig.free}
\end{figure}

Fig. \ref{fig.compare} presents qualitative comparisons with existing methods. We reconstruct instances from the input image (shown in the first column) and render two novel views of the object, with the ground truth images displayed in the last two columns. For each category, we show the reconstruction results for two instances. Existing methods, which rely on local feature retrieval, often produce blurry and distorted results when the viewpoint deviates from the input image, as seen in the Hydrant and the first Cup samples. In contrast, our method utilizes global multi-level features, resulting in sharper and more 3D-consistent reconstructions. Furthermore, our method benefits from the category-specific template and foreground loss, enabling clearer boundaries between the foreground and background in novel views. Other methods, such as GRF, often generate novel views with unclear backgrounds, as shown in the first Mouse sample. We also find that methods like Henzler \etal, NerFormer, and VisionNeRF struggle with non-Lambertian surfaces. Specifically, VisionNeRF severely fails to render plausible novel views for objects in the Mouse and Cellphone categories, which feature reflective surfaces. Our method, on the other hand, demonstrates greater robustness by correctly reconstructing these objects.

To further demonstrate the 3D-consistent reconstruction of our method, we render six novel views from random viewpoints for objects from each category and present the qualitative results in Fig. \ref{fig.free}. Our method not only produces high-quality novel views with clear foreground-background boundaries, but also maintains 3D consistency across all viewpoints, which demonstrates the robustness of our method to viewpoint variations from the input images.

\subsection{Ablation Study}

\subsubsection{Effectiveness of each Component}

\begin{table}
    \centering
    \small
    \caption{Ablation study of each component. The "Template" column indicates whether the category-specific shape template is used. The "Alignment" column shows whether each instance is aligned with the template. The "Multi-Scale Feature" column compares the use of multi-scale fused features versus the use of only the last scale feature in the object encoding module. The "Pre-Training" column indicates whether the proposed contrastive learning-based pre-training method is employed.}
    \label{tab.ablation}
    \resizebox{0.98\linewidth}{!}{
        \begin{tabular}{c||cccc||ccc}
            \hline
            Method      & Template     & Alignment    & Multi-Scale Feature & Pre-Training & PSNR $\uparrow$ & SSIM $\uparrow$ & LPIPS $\downarrow$ \\ \hline
            Baseline    &              &              &                     &              & 20.274          & 0.817          & 0.209             \\
            Variant \#1  & $\checkmark$ &              &                     &              & 19.349          & 0.808           & 0.209              \\
            Variant \#2  &              & $\checkmark$ &                     &              & 21.532          & 0.831           & 0.182              \\
            Variant \#3  & $\checkmark$ & $\checkmark$ &                     &              & 22.258          & 0.854           & 0.174              \\
            Variant \#4  & $\checkmark$ & $\checkmark$ & $\checkmark$        &              & 22.480         & 0.858           & 0.169              \\
            Variant \#5  & $\checkmark$ & $\checkmark$ &                     & $\checkmark$ & {\ul 22.622}    & {\ul 0.859}     & {\ul 0.169}        \\
            \hline
            Full Method & $\checkmark$ & $\checkmark$ & $\checkmark$        & $\checkmark$ & \textbf{22.987} & \textbf{0.861}  & \textbf{0.168}     \\ \hline
        \end{tabular}
    }
\end{table}

We evaluate the effectiveness of each component in our framework through an ablation study on the Banana category, with the quantitative results reported in Tab. \ref{tab.ablation}. We begin with a baseline method that omits all the proposed components. In this baseline, instances are not aligned with the template, and the density $\sigma$ of $x$ is predicted directly without the use of the shape template. Additionally, only the feature from the last layer of the feature extractor $f_e$ is used.

Variant \#1 introduces the shape template to the baseline but does not improve performance compared to the baseline. Variant \#2 adds instance alignment to the baseline, resulting in a performance improvement, highlighting the importance of aligning instances to the template. However, the performance of Variant \#2 is still inferior to Variant \#3, which incorporates both the shape template and instance alignment, demonstrating that both components are crucial for improved reconstruction.

Variant \#4 further introduces multi-scale features for object encoding, which outperforms Variant \#3 that uses only the last scale feature. This demonstrates the benefits of multi-scale feature fusion for capturing richer object representations. Finally, Variant \#5 incorporates the contrastive learning-based pre-training method, leading to a substantial improvement over Variant \#3. When all components are combined, the full method achieves the best performance, demonstrating the effectiveness of VRF.

\subsection{Effectiveness of Instance-Specific Rendering}

\begin{table}
    \centering
    \caption{Quantitative comparison between instance-specific method and category-specific methods of various model size. Experiment is conducted on Hydrant category.}
    \label{tab.instance}
    \resizebox{0.8\linewidth}{!}{
        \begin{tabular}{c||c||c|ccc}
            \hline
            Method                             & Hidden Neurons & Sec/Frame     & PSNR $\uparrow$ & SSIM $\uparrow$ & LPIPS $\downarrow$ \\ \hline
            \multirow{3}{*}{Category-Specific} & 192        & {\ul 0.92}    & 17.784          & 0.825           & 0.177              \\
            & 512        & 2.06          & 18.392          & 0.838           & 0.168              \\
            & 1280       & 7.76          & {\ul 18.783}    & {\ul 0.841}     & {\ul 0.166}        \\ \hline
            Instance-Specific                  & 192        & \textbf{0.75} & \textbf{19.009} & \textbf{0.847}  & \textbf{0.162}     \\ \hline
        \end{tabular}
    }
\end{table}

\begin{table}
    \centering
    \small
    \caption{Ablation study on initial template selection. We randomly select four instances from the Hydrant training data as the initial template and compare their performance.}
    \label{tab.template}
    \resizebox{\linewidth}{!}{
        \begin{tabular}{c||cccc}
            \hline
            Metric             & 427\_60021\_116224 & 106\_12648\_23157 & 194\_20953\_43984 & 342\_35797\_64722 \\ \hline
            PSNR $\uparrow$    & 19.009             & 18.688          & 18.982          & 18.977          \\
            SSIM $\uparrow$    & 0.847              & 0.839           & 0.846           & 0.843           \\
            LPIPS $\downarrow$ & 0.162              & 0.170           & 0.165           & 0.165           \\ \hline
        \end{tabular}
    }
\end{table}

Our instance creation module utilizes a hyper-network-based strategy for efficient instance-specific rendering. To demonstrate its effectiveness, we conduct an ablation study using category-specific neural networks with varying numbers of hidden neurons to predict the density and color of points in the radiance field during rendering. The quantitative results are presented in Table \ref{tab.instance}. The category-specific method requires high-dimensional appearance and geometry representations $z_d$ and $z_c$ as input, while the instance-specific method only requires point positions. As a result, the category-specific method with the same number of hidden neurons is slower than its instance-specific counterpart. Increasing the number of hidden neurons significantly improves the performance of the category-specific method but also decreases the rendering speed. Compared to the category-specific method with 1280 hidden neurons, our instance-specific method achieves a $1.2\%$ improvement in PSNR while rendering an image $90.3\%$ faster. This highlights the effectiveness of our hyper-network-based rendering strategy.

\subsubsection{Robustness to Initial Template Selection}

\begin{figure}
    \centering
    \begin{minipage}{0.48\textwidth}
        \centering
        \includegraphics[width=\textwidth]{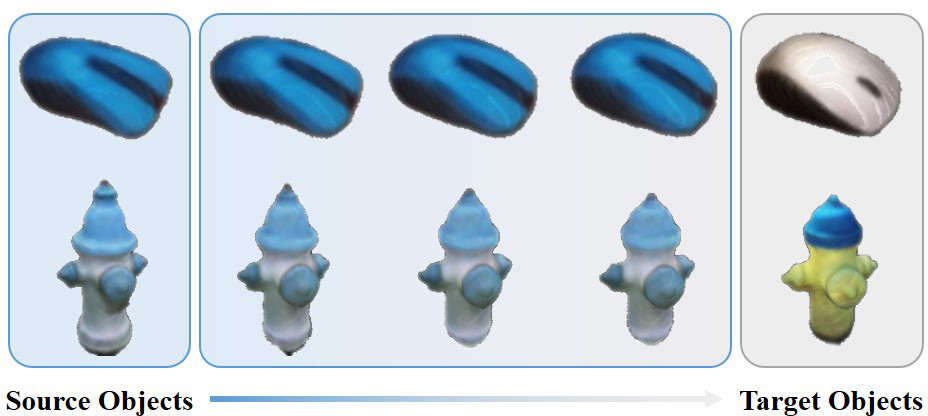}
        \caption{We linearly mix the geometry representation $z_d$ of the source and target objects, leading to gradually evolving geometries.}
        \label{fig.inter}
    \end{minipage}
    \hfill
    \begin{minipage}{0.46\textwidth}
        \centering
        \includegraphics[width=\textwidth]{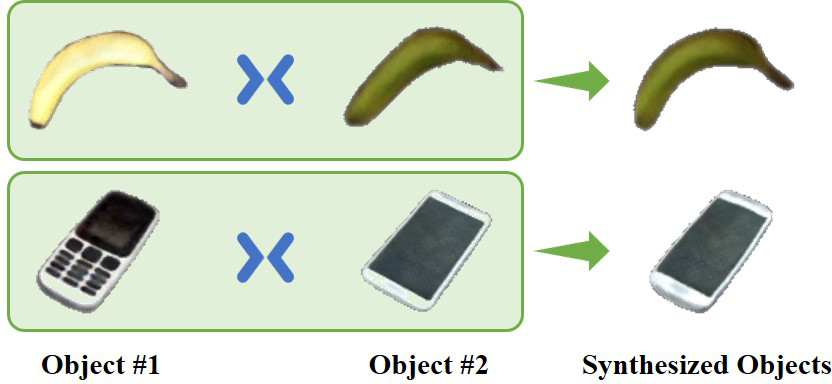}
        \caption{We create a new object with the geometry representations $z_d$ from Object \#1 and the appearance representations $z_c$ from Object \#2.}
        \label{fig.new}
    \end{minipage}
\end{figure}

The category-specific shape template $f_t$ is initialized by optimizing a vanilla NeRF using an instance randomly selected from the training set. For the Hydrant category, we choose the instance numbered \emph{427\_60021\_116224} to optimize the initial shape template. To demonstrate the robustness of our method to the choice of the initial template, we also experiment with three other instances from the Hydrant category training set. The quantitative results are reported in Tab. \ref{tab.template}. The final performance across all four models is similar, confirming that our method is robust to the selection of the initial template. Notably, the instance numbered \emph{106\_12648\_23157}, which exhibits significant shape deviation from a typical hydrant, also yields comparable performance when used as the initial template.

\subsubsection{Global Representation Property}

Our method leverages two distinct global representations to parameterize the appearance and geometry of an object. We demonstrate that these global representations exhibit good continuity in high-dimensional space, as shown in Fig. \ref{fig.inter}. Specifically, we linearly mix the geometry representations $z_d$ of the source and target objects, while maintaining the appearance representation $z_c$ of the source object. This allows us to create a new object whose geometry transitions smoothly from the source object to the target object. In Fig. \ref{fig.new}, we demonstrate the decomposed geometry and appearance representation by creating new objects with the geometry representation $z_d$ from Object \#1 and the appearance representation $z_c$ from Object \#2. The created object retains the geometry of Object \#1 and the appearance of Object \#2, highlighting the distinct roles of geometry and appearance in VRF.

\section{Limitation and Future Work}

Our method is currently based on the original NeRF, which is constrained by slow rendering speeds. While we have introduced a hyper-network-based acceleration strategy and demonstrated its effectiveness, further improvements in rendering speed would significantly enhance real-world applicability. Therefore, we plan to integrate advanced techniques, such as instant-ngp \cite{instant-ngp} and 3D gaussian splatting \cite{3d_gs}, in future work to further optimize our method. Additionally, although our current focus is on object reconstruction, our method is also well-suited for human reconstruction. We intend to explore human reconstruction using our approach in future research.

\section{Conclusion}

We propose a framework, named VRF, that efficiently reconstructs category-specific objects from a single image. Our framework leverages global representations to separately parameterize the geometry and appearance of an object, eliminating the need for frequent and fragile projection-based feature retrievals. To address the coordinate inconsistency problem inherent in real-world data, we define a canonical space by introducing a category-specific template. This approach not only mitigates alignment issues but also reduces the complexity of geometry modeling by deforming the template to represent different instances. Additionally, we introduce a hyper-network-based NeRF creation scheme for efficient rendering, complemented by a contrastive learning-based pretraining strategy to enhance model performance. As a result, our framework surpasses existing methods in both reconstruction quality and rendering speed.

\bibliographystyle{elsarticle-num}
\bibliography{references}
\end{document}